\RequirePackage{amsmath}
\documentclass[runningheads]{llncs}
\usepackage[T1]{fontenc}
\usepackage{graphicx}
\newcommand{\E}{\mathbb{E}} %

\usepackage[dvipsnames]{xcolor}  %
\usepackage{amssymb} %
\usepackage{mathtools} %
\usepackage{hyperref} %
\usepackage[capitalize,noabbrev,nameinlink]{cleveref}
\usepackage{tikz}  %
\usepackage{graphicx} %
\usepackage{listings}

\graphicspath{{img/}}
\Crefname{equation}{Eq.}{Eqs.}
\Crefname{figure}{Fig.}{Figs.}
\Crefname{tabular}{Tab.}{Tabs.}
\Crefname{table}{Tab.}{Tabs.}
\usepackage{color}

\usepackage{eso-pic}
\begin{document}
\AddToShipoutPictureFG*{\AtPageLowerLeft{\put(\LenToUnit{\dimexpr(\paperwidth-\textwidth)/2\relax},\LenToUnit{0.45in}){\parbox{\textwidth}{\footnotesize\centering Preprint (version submitted before peer review). Published in: Discovery Science (DS 2024), Lecture Notes in Computer Science, vol.~15244, Springer, \url{https://doi.org/10.1007/978-3-031-78980-9_18}}}}}
\title{Pairwise Difference Learning for Classification}
\titlerunning{Pairwise Difference Learning for Classification}
\author{Mohamed Karim Belaid\inst{1,2}\orcidID{0000-0001-9038-9045} \and
Maximilian Rabus\inst{2}\orcidID{0000-0003-0755-1772} \and
Eyke H\"ullermeier\inst{3}\orcidID{0000-0002-9944-4108}}
\authorrunning{M. Belaid et al.}
\institute{IDIADA Fahrzeugtechnik GmbH, Munich, Germany  
\email{karim.belaid@idiada.com}\\
\and
Dr.\ Ing.\ h.c.\ F.\ Porsche AG Stuttgart, Germany
\email{\{extern.karim.belaid,maximilian.rabus2\}@porsche.de}\\
\and
Ludwig-Maximilians-Universit\"at  Munich, Germany\\
\email{eyke@ifi.lmu.de}}

\maketitle              

\begin{abstract}
Pairwise difference learning (PDL) has recently been introduced as a new meta-learning technique for regression. Instead of learning a mapping from instances to outcomes in the standard way, the key idea is to learn a function that takes two instances as input and predicts the difference between the respective outcomes. Given a function of this kind, predictions for a query instance are derived from every training example and then averaged. This paper extends PDL toward the task of classification and proposes a meta-learning technique for inducing a PDL classifier by solving a suitably defined (binary) classification problem on a paired version of the original training data. We analyze the performance of the PDL classifier in a large-scale empirical study and find that it outperforms state-of-the-art methods in terms of prediction performance. Last but not least, we provide an easy-to-use and publicly available implementation of PDL in a Python package. 

\end{abstract}

\section{Introduction}
Pairwise difference learning (PDL) has recently been introduced independently by Tynes et al.\ \cite{tynes2021pairwise} and Wetzel et al.\ \cite{wetzel2022twin} as a meta-learning technique for regression, which transforms the original task of learning to predict outcomes for individual inputs into the task of learning to predict \emph{differences} between the outcomes of input \emph{pairs}: 
Noting that the value of a function $f$ at a point $x$ can be written ``from the perspective'' of any other point $x'$ as $f(x) = f(x') + \Delta (x,x')$ with $\Delta(x,x') = f(x) - f(x')$, the simple idea of PDL is to train an approximation $\tilde{\Delta}$ of the difference function $\Delta$ and obtain predictions of new outcomes $y = f(x)$ by averaging over the predicted differences to the outcomes in the training data:
\begin{equation}\label{eq:simple}
y  \approx \frac{1}{N} \sum_{i=1}^N y_i + \tilde{\Delta}(x,x_i)
\end{equation}
One of the main motivations of PDL is the quadratic increase of the training data: If the original training data contains $N$ data points $(x_1, y_1), \ldots , (x_N, y_N)$, the difference function can be trained on potentially $\mathcal{O}(N^2)$ training examples of the form $((x_i, x_j), y_i - y_j)$. This increase might be specifically useful in the ``small data'' regime (even if the transformed examples are of course no longer independent of each other). Moreover, note that the prediction (\ref{eq:simple}) benefits from a statistically useful averaging effect.

Building on the basic idea of PDL, we make the following contributions. %
We extend the idea of PDL toward the task of classification and propose the PDL classifier, a meta-learning approach that transforms any multiclass classification problem into a single binary problem. This innovative method leverages the concept of learning inter-class differences, leading to demonstrably improved average prediction accuracy (\cref{sec:method_multiclass}). 
We introduce the ``pairwise difference learning library'' (pdll) on PyPI, which incorporates our implementation of the PDL classifier and ensures compatibility with any Sklearn ML model (\cref{sec:method_library}). 
We conduct a large-scale experimental analysis of PDL and compare the results to state-of-the-art ML estimators (\cref{sec:eval}). 
We discuss the architecture of PDL and how it can lead to an improvement of the accuracy (\cref{sec:discussion}).

\section{Related Work}
Tynes et al. introduced pairwise difference regressor \cite{tynes2021pairwise}, a novel meta-learner for chemical tasks that enhances prediction performance, compared to random forest and provides robust uncertainty quantification. In computational chemistry, estimating differences between data points helps mitigate systematic errors \cite{tynes2021pairwise}.
In parallel, Wetzel et al. used twin neural network architectures for semi-supervised regression tasks, focusing on predicting differences between target values of distinct data points \cite{wetzel2022twin}. The approach of Wetzel et al. enabled training on unlabelled data points when paired with labeled anchor data points. By ensembling predicted differences between target values, the method achieved high prediction performance for regression problems. While conceptually similar to the pairwise difference regressor in emphasizing differences between data points, it is specialized to neural network architectures for semi-supervised regression tasks~\cite{wetzel2022btwin}.

The pairwise difference learning (PDL) literature has since then, evolved into diverse methodologies and applications. Spiers et al. measured sample similarity in chemistry, emphasizing spectral shape differences using metrics like Euclidean and Mahalanobis distances. They extended the approach by calculating a Z-score which offers insights into prediction accuracy, facilitating outlier detection and model adaptation \cite{spiers2023physicochemical}.
PDL was developed mainly for regression tasks. It can also be adapted to targets that might be known or only bounded. Example of target annotations could be $y=5.3$, $y<2.1$, or $y>6.5$. Predicting an increase/decrease between a pair is a possible solution \cite{fralish2024leveraging}. 
PDL regressor with its variants has demonstrated efficacy in various applications, including regression with image input \cite{hu2023exploring}, learning chemical properties \cite{fralish2023deepdelta}, quantum mechanical reactions \cite{chen2023benchmark}, and drug activity ranking \cite{wang2023extrapolation}. %

\section{PDL Classification}\label{sec:method_multiclass}

Consider a standard setting of supervised (classification) learning: Given a set of training data 
$$
\mathcal{D} = \{ (x_i , y_i ) \}_{i=1}^N \subset \mathbb{R}^d \times \mathcal{Y} \, ,
$$
comprised of training instances in the form of feature vectors $x_i \in \mathbb{R}^d$ together with observed discrete labels $y \in \mathcal{Y} = \{ 1, \ldots, K \}$, and assumed to be generated i.i.d.\ according to an underlying (unknown) joint probability measure $P$, the task is to learn a predictor $\textup{PDC}: \mathbb{R}^d \rightarrow \mathcal{Y}$ with low risk (expected loss). 
The PDL classifier transforms the original training data $\mathcal{D}$ into the new data
\begin{equation}\label{eq:ttd}
\mathcal{D}_{pair} = \big\{ (z_{i,j} , y_{i,j}) \, \vert \, 1 \leq i , j \leq N \big\} \, ,
\end{equation}
where $z_{i,j} = \phi(x_i,x_j)$ is a \emph{joint} feature representation of the instance pair $(x_i,x_j)$ and 
\begin{equation}\label{eq:Kronecker_delta}
y_{i,j}  = \begin{cases}
0 & \text{for } y_i \neq y_j, \\
1 & \text{for } y_i = y_j 
\end{cases} \, .
\end{equation}
Thus, we seek a binary classifier $\gamma: \, \mathbb{R}^{d} \times \mathbb{R}^{d} \rightarrow [0, 1 ]$ that, given two instances $x$ and $x'$ as input, predicts whether or not the respective classes $y$ and $y'$ are the same. 
More specifically, we assume $\gamma$ to be a probabilistic classifier, so that $\gamma(x,x') \in [0,1]$ is the probability that $y = y'$. Deterministic classifiers that return a binary label as a prediction are treated as degenerate $\{0,1\}$-valued probabilistic classifiers. 
Leveraging the joint feature representation, $\gamma$ is of the form $\gamma(x,x') = h(\phi(x,x'))$, where $h$ is trained on the transformed data (\ref{eq:ttd}). To this end, any binary classification method can be used.  
Note, however, that the binary problem might be quite imbalanced, as the transformation (\ref{eq:Kronecker_delta}) will produce much more negative (unequal) than positive (equal) examples. One can solve this issue by introducing class weights \cite{king2001logistic} to equalize the loss function of the classifier $\gamma$. 
As for the joint feature representation, the original proposal was to define $z_{i,j}$ as a concatenation of $x_i$ and $x_j$. It turned out, however, that expanding this vector by the difference $x_i - x_j$ has a positive influence on performance \cite{tynes2021pairwise}, wherefore we also adopted this representation in our work. 

\begin{figure}
    \centering
    \includegraphics[width=\linewidth]{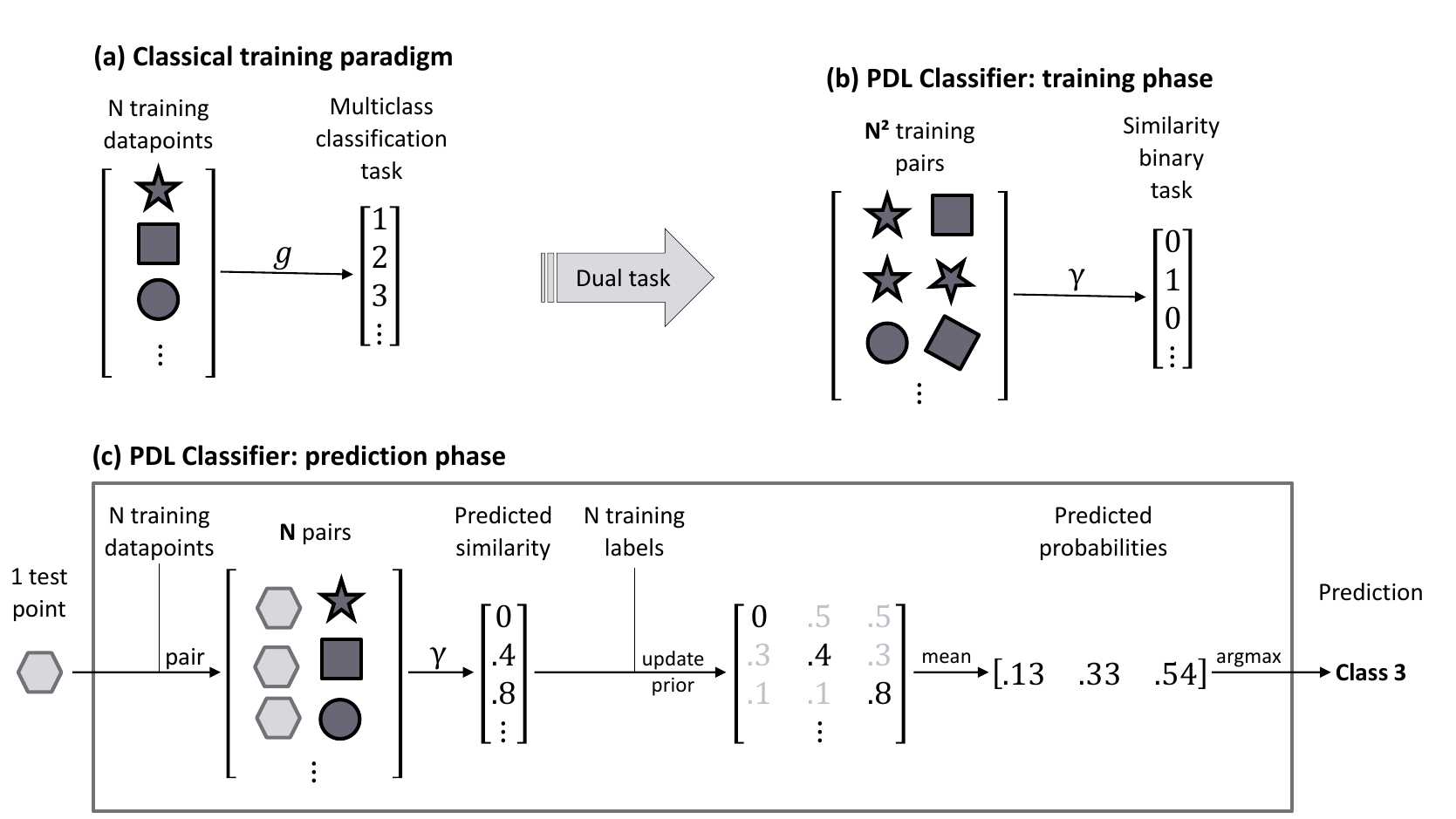}
    \caption{Illustration of the PDL classifier.}
    \label{fig:pdc_graphical_abstract}
\end{figure}

Since (class) equality is a symmetric relation, $\gamma$ is naturally expected to be symmetric in the sense that $\gamma(x_i,x_j) = \gamma(x_j,x_i)$. By adding both $(\phi(x_i,x_j), y_{i,j})$ and $(\phi(x_j,x_i), y_{j,i})$ to $\mathcal{D}_{pair}$, this symmetry can also be reflected in the training data. But even then, however, $\gamma$ is not necessarily guaranteed to preserve symmetry. Therefore, we additionally ``symmetrize'' the predictor as follows:
\begin{equation}\label{eq:multiclassifier_predict_symmetry}
    \gamma_{sym}(x_i , x_j) = \frac{\gamma(x_i , x_j) + \gamma(x_j , x_i)}{2}
\end{equation}
Given a query $x_q$, we finally estimate the probability of class labels $y \in \mathcal{Y}$ as follows: Considering each training example $(x_i, y_i)$ as a piece of evidence for the unknown class $y_q$, the semantics of the above prediction suggests that the probability of the event $y_q = y_i$ is given by (\ref{eq:multiclassifier_predict_symmetry}). More formally, $P(E) = \gamma_{sym}(x_q , x_i)$, where $E$ denotes the event $y_q = y_i$ (and hence $P(\neg E) = 1 - \gamma_{sym}(x_q , x_i)$). Let $p$ denote the prior distribution on the class labels $\mathcal{Y}$ (which can easily be estimated by relative frequencies on the training data). This distribution is then updated by conditioning it on the (uncertain) event $E$, which yields the following posterior suggested by $(x_i, y_i)$:
\begin{equation}
p_{post,i} (y) = \left\{ \begin{array}{cl}
 \gamma_{sym}(x_q , x_i) & \text{ if } y = y_i \\[2mm]
 \dfrac{p(y) \cdot (1 - \gamma_{sym}(x_q , x_i))}{1- p(y_i)} & \text{ otherwise}
\end{array} \right.
\end{equation}
Thus, the (posterior) probability of $y_i$ is fixed to $\gamma_{sym}(x_q , x_i)$, and all other probabilities are rescaled in a proportional way, to guarantee that the sum of posterior probabilities adds to 1. 
Finally, we average over the evidences from all training examples to obtain 
\begin{equation}\label{eq:bma}
p_{post}(y) = \frac{1}{N} \sum_{i=1}^N p_{post,i} (y) \, .
\end{equation}
In case a deterministic prediction is sought, the class with the highest (estimated) probability is chosen: 
\begin{equation}\label{eq:multiclassifier_predict}
\hat{y}_q = \arg \max_{y \in \mathcal{Y}}  p_{post}(y)
\end{equation}

\subsection{Uncertainty Quantification}\label{sec:uq}

Interestingly, the PDL approach also offers a natural approach to uncertainty quantification, a topic that has received increasing attention in the recent machine learning literature. In particular, recent research has focused on the distinction between so-called \emph{aleatoric} uncertainty (caused by inherent randomness in the data) and \emph{epistemic} uncertainty (caused by the learner's incomplete knowledge of the true data-generating process)\,---\,we refer to \cite{mpub440} for a detailed exposition of this topic. 

Within the Bayesian approach, these two types of uncertainty can be captured by properties of the posterior predictive distribution, which in turn can be approximated through ensemble learning \cite{laks_sa17}. In a sense, PDL parallels this approach, with each anchor playing the role of an ensemble member, and (\ref{eq:bma}) mimicking Bayesian model averaging. This suggests the following quantification of aleatoric (AU), epistemic (EU), and total uncertainty (TU) of a prediction, with $H$ denoting Shannon entropy.:
\begin{align*}
    \text{TU} & = H(p_{post}(y)) = H \left( \sum_{i=1}^N p_{post,i} (y)  \right) \\
    \text{AU} & = \frac{1}{N} \sum_{i=1}^N H ( p_{post,i} (y) ) \\
    \text{EU} & = \text{TU} - \text{AU} \\
\end{align*}
Theoretically, these measures are justified based on a well-known result from information theory, according to which entropy additively decomposes into conditional entropy and mutual information \cite{depe_du18}. Broadly speaking, the more uniform the (averaged) distribution $p_{post}$, the higher the total uncertainty, and the more diverse the individual predictions $p_{post,i}$, the higher the epistemic uncertainty.

\subsection{Illustration}
Thanks to its novel structure, the PDL classifier can solve a multiclass classification task by training exactly one instance of a base learner on a binary task. 
\cref{fig:pdc_graphical_abstract} illustrates the PDL classifier algorithm, showcasing both the training and prediction phases on a simple multiclass task. 
\cref{fig:pdc_graphical_abstract}.a shows a traditional multiclass classifier $g$ that maps each of the N training data points to their assigned unique class label (star, square, or circle). 
In \cref{fig:pdc_graphical_abstract}.b, PDL classifier transforms the data by creating $N^2$ pairs of data points. During training, a binary classifier $\gamma$ learns to distinguish between pairs that belong to the same class (positive label) from pairs of different classes (negative label).
In \cref{fig:pdc_graphical_abstract}.c, given one query input, the PDL classifier pairs it with each of the N training data points. For each pair, the classifier predicts a probability of similarity (belonging to the same class). Predicted probabilities are mapped to the column corresponding to the initial label of each training point. Missing posterior probabilities, in grey, are estimated by updating the prior probabilities, assuming a uniform distribution in this example. Finally, averaging across all training points yields the predicted probabilities for each class. The class with the highest predicted probability is chosen as the final class label for the query point (e.g., Class 3).

\begin{figure*}
    \centering
    \includegraphics[width=\linewidth]{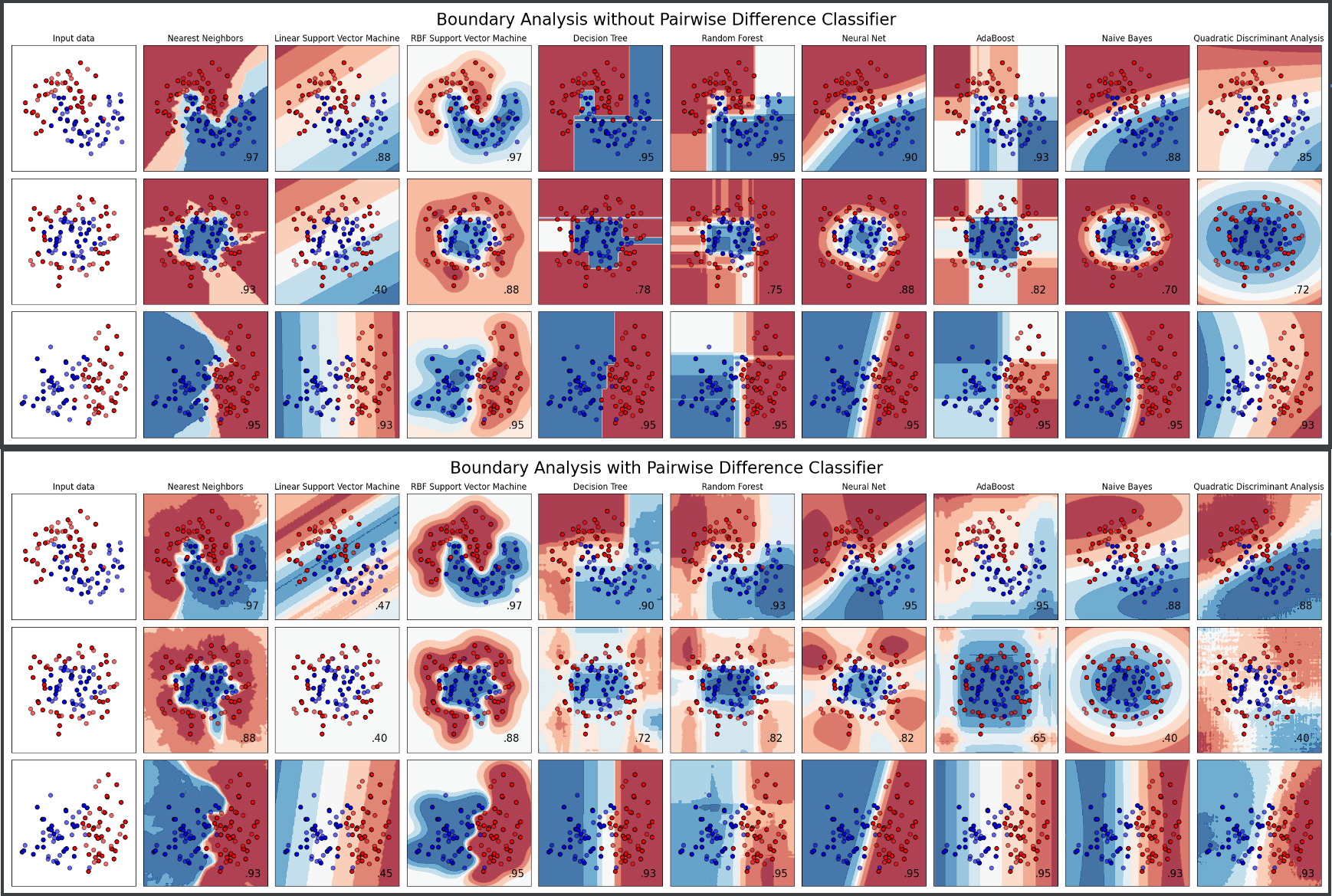}
    \caption{Comparing learned patterns using PDL classifiers and baseline models.}
    \label{fig:2d_datasets}
\end{figure*}
\cref{fig:2d_datasets} illustrates the patterns learned by nine baseline models across three 2D datasets. The baseline 3-Nearest-Neighbor (3-NN) classifier can only predict four probabilities: $0, \frac{1}{3}, \frac{2}{3},$ and $1$. This is evident in the figure, where each dataset shows only four discrete regions. In contrast, when using PDL on top of 3-NN, the predicted probability is derived from the averaging over $N$ discrete predictions. This results in more refined and precise probability estimates. 
Despite the simplicity of some estimators, PDL leverages more complex patterns. The contrast between DecisionTree with and without PDL clearly illustrates PDL's capability to learn non-linear patterns. 
The underfitting observed when incompatible base models learn corrupted patterns underscores the critical role of the choice of base learners.

\subsection{Choice of Base Learners}\label{sec:method_compatibility}

As already said, PDL can theoretically be implemented with any (probabilistic) binary classifier as a base learner\,---\, or, stated differently, it can be used as a wrapper for any (binary or multinomial) classifier. Practically, however, some classifiers might be more suitable as base learners and others less. 

One thing one should keep in mind is that even if the original data $\mathcal{D}$ is i.i.d., independence will be lost for $\mathcal{D}_{pair}$ as soon as the same instance $x_i$ is paired with various other instances. This is very similar to the setting of metric learning, where models are also trained on pairs of data points \cite{bian_la13}. In practice, although many machine learning algorithms turn out to be quite robust against violations of the i.i.d.\ assumption \cite{kutner2005applied}, some methods may be concerned more than others. 

Another important aspect is the joint feature representation $ z = \phi(x,x')$. For example, by defining $z$ as a concatenation of $x$, $x'$, and the difference $x-x'$, one obviously introduces (perfect) multicollinearity. Again, while this is problematic for some machine learning methods, notably linear models \cite[p.8]{tynes2021pairwise}, others can deal with this property more easily. 

While an in-depth analysis of the suitability of different base learners is beyond the scope of this paper, we generally found that non-parametric methods are more robust and tend to show better performance than parametric ones. In our experimental evaluation, we will therefore mainly use tree-based methods, which have the additional advantage of being fast to train.   

\subsection{Complexity}
Looking at the complexity of PDL, suppose the complexity of a base learner to be $\mathcal{O}(p(N, M, F, K))$, where $p(\cdot)$ is polynomial in the number of training points ($N$), 
the number of test points ($M$),
the number of input features ($F$), and 
the number of output classes ($K$). %
The complexity of PDL is then $\mathcal{O}(p(N^2,2M A ,3F,2))$:
The training points are scaled to $N^2$ pairs; 
the features are scaled to $3F$ ($F$ features of point $x_i$, $F$ features of point $x_j$, and $F$ features of the difference $x_i-x_j$. This feature construction technique for PDL has demonstrated previously improved results \cite{tynes2021pairwise});
Each test point is paired with the $A$ anchor points. Pairs are duplicated twice to obtain their symmetry.
Thus, $M$ test predictions of PDL require $2MA$ predictions using the base learner.
The number of output classes $K$ shrinks to $2$ since the model is asked to predict whether the pair of points has a similar class.

\subsection{PDL Library}\label{sec:method_library}
Our library\footnote{Link: \url{https://github.com/Karim-53/pdll}} includes a Python implementation of the PDL classifier, adhering to the Scikit-learn standards. Consequently, integrating the PDL classifier into existing codebases is straightforward, requiring minimal modifications. As demonstrated in the example below, only two additional lines of code are needed:
\centerline{\includegraphics[width=\columnwidth]{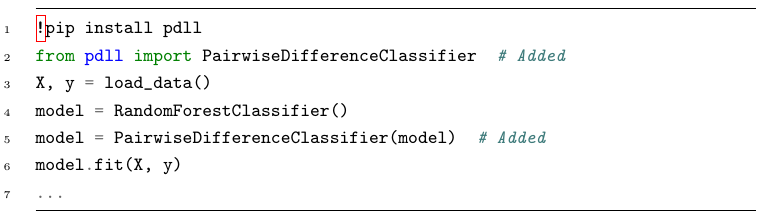}}

\section{Evaluation}\label{sec:eval}
In this section, we test PDC on various public datasets from OpenML \cite{OpenML2013} and compare it to 7 Scikit-learn state-of-the-art learners. 

\subsection{Data}

\begin{figure}[htb]
\centerline{\includegraphics[width=\columnwidth]{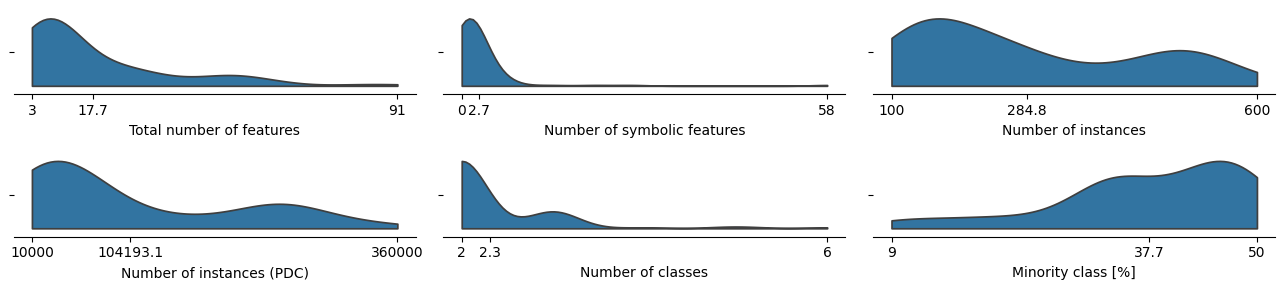}}
\caption{Distribution of key characteristics of the 99 OpenML classification datasets (minimum, mean, maximum).} \label{fig:describe_data_exp2}
\end{figure}

OpenML provides a diverse range of datasets, many of which are small, with 37\% having less than 600 data points. This study focuses on small datasets, for which the pairwise learning approach is presumably most effective.
We applied grid search CV for parameter tuning, leveraging the search space from TPOT \cite{olson2016evaluation}. To accommodate our grid search setup, we subsampled the search space to 1,000 parameter combinations per estimator. %
Following dataset selection constraints similar to the OpenML-CC18 benchmark \cite{bischl2017openml}, we randomly selected 99 datasets (see summary statistics in \cref{fig:describe_data_exp2}). Although these datasets are relatively small, the effective data size for PDC is quadrupled due to the pairing, reaching $360 000$ data points. We also monitored class imbalance using the ``minority class'' meta-data, which represents the percentage of the minority class relative to the total size of each dataset.
Considering the 7 baseline models, we performed 5 times 5-fold CV with an inner 3-fold grid search CV, totaling 66~528~000 train-test runs and 3 weeks wall-time on an HPC. %

\subsection{Data Processing Pipeline}

Using scikit-learn \cite{pedregosa2011scikit}, we implemented a common data processing pipeline for all runs,
with standardization for numeric features, one-hot encoding for nominal features, and ordinal encoding for ordinal features. Since PDL needs the pair difference $x_i - x_j$ as additional inputs, processed features are all treated as numeric when applying the difference. %

\subsection{Performance Measures} 

We measure performance in terms of the (macro) F1 score, which is arguably more meaningful than the standard misclassification rate in the case of imbalanced data. In binary classification, the F1 score is defined as the harmonic mean of precision and recall. For multinomial problems, the macro version of this score is the (unweighted) mean of the F1 scores for the individual class:
$$
\text{Macro} F1 = \frac{1}{K} \sum_{i=1}^{K} F1_i \, ,
$$
where $F1_i$ is the F1 score on the $i^{th}$ class (treating test examples of this class as positive and all others as negative). 
We also report the improvement of PDL over the base learner in terms of the difference
$\Delta F1 = \text{Macro} F1_{PDC} - \text{Macro} F1_{base}$.
We aggregate the results using the mean $\pm$ standard error. 

To aggregate the results of all data sets, we count the number of wins/losses by comparing the average performance of models over 25 runs (5 times 5-fold CV) per dataset. A win is counted when PDC's average score is higher than the baseline; a loss is counted otherwise. 
To determine the number of significant wins/losses, a Student's t-test is conducted for each dataset to assess the statistical significance of the difference in performance. A significant win/loss is recorded when the p-value of the t-test is below a predetermined threshold $\alpha=0.05$.
In some cases, there may be a tie in the average scores, leading to instances where the number of wins and losses does not sum to 99, which is the total number of datasets benchmarked. 

As an alternative to counting wins and losses, and despite being aware of the questionable nature of this statistic, we also average performance over data sets. Average performance may provide a first overall impression, although we agree that it should always be interpreted in a cautious way.  

\subsection{Results}\label{sec:results}

\begin{figure}[htb]
\centerline{\includegraphics[width=\columnwidth]{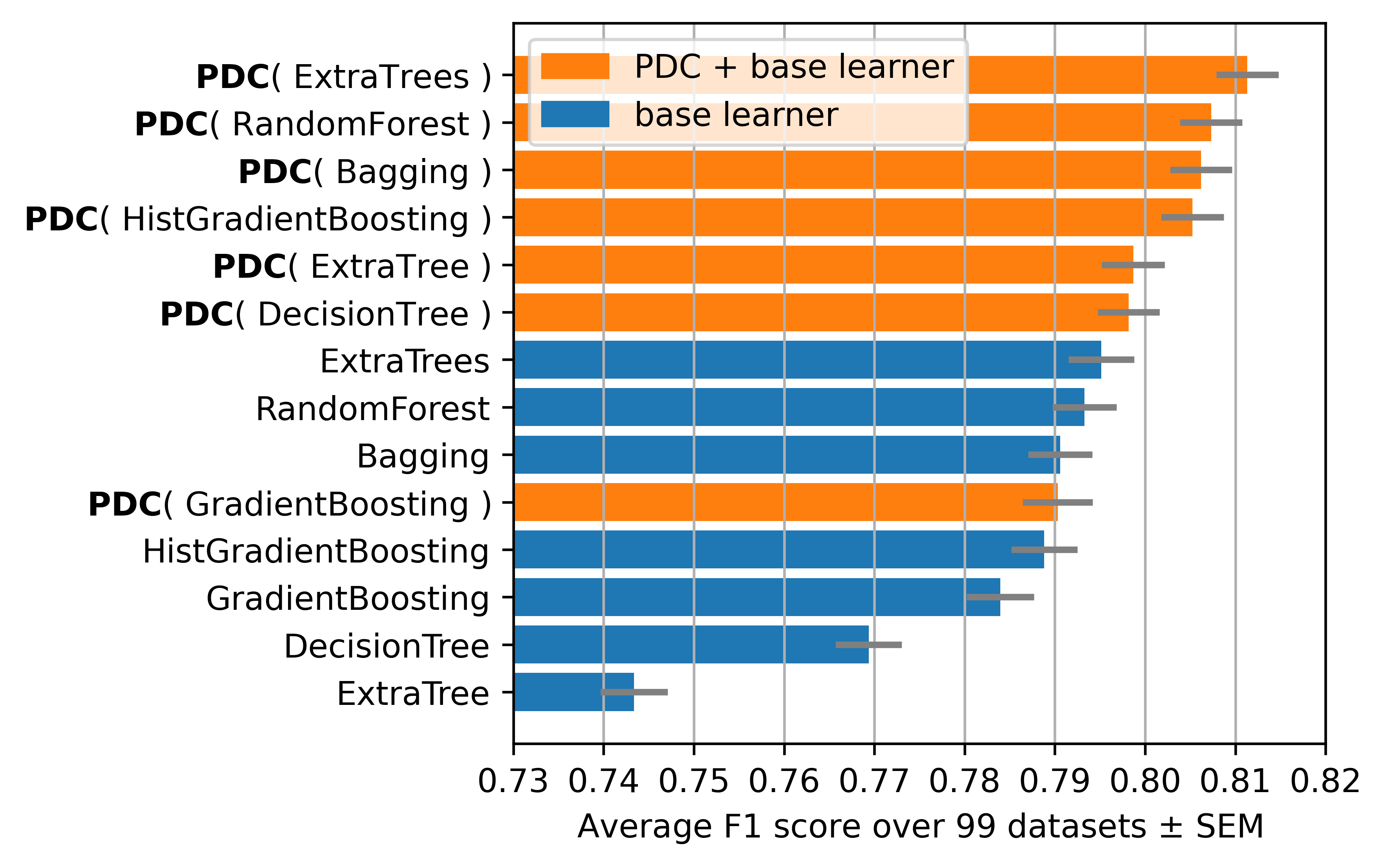}}
\caption{Comparing average Macro F1 score of optimized baseline classifiers and PDL classifiers.} \label{fig:exp2_bar_f1_classification}
\end{figure}

First, the PDL classifier, on top of ExtraTrees, obtained the best average Macro F1 score over the 99 datasets, outperforming all baselines, see \cref{fig:exp2_bar_f1_classification}.
In \cref{tab:exp2_f1_wins}, the ratio of significant wins demonstrates an advantage for the PDL classifier, suggesting that, in a one-to-one comparison, PDL is more likely to outperform its equivalent baseline.

\begin{table}
\caption{Comparing baseline classifiers to PDC using 99 datasets.}\label{tab:exp2_f1_wins}
\begin{center}
\begin{tabular}{lrr|rr|rrrr}
 & \multicolumn{2}{c|}{\textbf{Significant wins}}  &\multicolumn{2}{c|}{\textbf{Wins}}  &\multicolumn{4}{c}{\textbf{Average Test Macro F1}} \\
Classifier & base & PDC & base & PDC & base & $\pm$ sem & PDC & $\pm$ sem \\
\hline 

Bagging & 3 & \textbf{26} & 27 & \textbf{70} & 0.7906 & 0.0035 & \textbf{0.8062} & 0.0034 \\
DecisionTree & 2 & \textbf{50} & 22 & \textbf{76} & 0.7694 & 0.0037 & \textbf{0.7982} & 0.0034 \\
ExtraTree & 1 & \textbf{61} & 9 & \textbf{90} & 0.7434 & 0.0037 & \textbf{0.7987} & 0.0035 \\
ExtraTrees & 6 & \textbf{24} & 21 & \textbf{77} & 0.7951 & 0.0036 & \textbf{0.8113} & 0.0035 \\
GradientBoosting & 9 & \textbf{23} & 25 & \textbf{72} & 0.7839 & 0.0037 & \textbf{0.7903} & 0.0039 \\
HistGradientBoosting & 2 & \textbf{32} & 15 & \textbf{82} & 0.7888 & 0.0037 & \textbf{0.8053} & 0.0035 \\
RandomForest & 5 & \textbf{27} & 22 & \textbf{73} & 0.7933 & 0.0035 & \textbf{0.8073} & 0.0034 \\

\hline 
\end{tabular}
\end{center}
\end{table}

The PDL classifier can be viewed as a method to simplify the trained model. As shown in \cref{fig:exp2_bar_f1_classification}, the test performance of PDC(DecisionTree) is equivalent to or better than that of the seven benchmarked state-of-the-art estimators. This indicates that, with the help of PDL, training a single tree can compete with ensemble methods that typically train around 100 trees. In this context, explaining a single tree may provide a more straightforward solution.

\paragraph{Analyzing the unique contribution.}
While PDL classifiers have high probabilities of outperforming baseline models in a one-to-one comparison, the ultimate goal of a data scientist is to obtain the best performance on each dataset. 
Before introducing PDC, the maximum achievable Macro F1 score was $0.8112 \pm 0.0035$ averaged over the 99 datasets. With the help of PDC, we achieve higher scores in 75 datasets, and the new record becomes $0.8243 \pm 0.0031$.
This advance showcases the unique contribution of PDC to the field of ML compared to existing algorithms. 
Moreover, PDC offers not only an important unique contribution to the record but also the highest contribution. Indeed, PDC's leave-one-out contribution to this record is $0.8243 - 0.8112 = 0.0131$ while popular estimators like HistGradientBoosting get no unique contribution, i.e., they are not able to outperform all other estimators on any of the 99 datasets, see \cref{tab:exp2_contirbution_f1}.
PDC's contribution is even 32 times more important than the best baseline.

\begin{table}[tbh]
\caption{Unique contribution of each estimator to the average Macro F1 score using the best optimized model on each dataset.} \label{tab:exp2_contirbution_f1}
\begin{center}
\begin{tabular}{lrr}
Estimator & \textbf{Unique contribution}& \textbf{Wins}\\
\hline 
ExtraTree & 0 & 0 \\
HistGradientBoosting & 0 & 0 \\
RandomForest & 0.00002 & 1 \\
Bagging & 0.00004 & 2 \\
GradientBoosting & 0.00006 & 2 \\
DecisionTree & 0.00020 & 10 \\
ExtraTrees & 0.00041 & 9 \\
\textbf{PDC} & \textbf{0.01312} & \textbf{75} \\
\hline 
\end{tabular}
\end{center}
\end{table}

\paragraph{Analyzing the overfitting.}
PDL classifiers have the advantage of decreasing overfitting. Indeed, looking at the 199 cross-validation (CV) runs in which both the baseline and PDL classifier obtain non-significant differences in train Macro F1 scores, we notice that PDL classifiers have a smaller train-test gap. 
A lower overfitting is observed when grouping by base classifier, see \cref{tab:overfit}. 
This even remains true without conditioning on non-significantly different train scores.

\begin{table}[tbh]
\caption{Comparing test Macro F1 on the subset of runs where train scores are not significantly different.} \label{tab:overfit}
\begin{center}
\begin{tabular}{lr|rr|rr|rr}
   &\textbf{\# CV} & \multicolumn{2}{c}{\textbf{Baseline Macro F1}}  & \multicolumn{2}{c}{\textbf{PDC Macro F1}} & \textbf{Test} & \textbf{Test}  \\
Estimator & \textbf{runs} & \textbf{Train} & \textbf{Test} & \textbf{Train} & \textbf{Test} &$\Delta F1$ & \textbf{p-value} \\
\hline 
Bagging & 20 & 0.998 & 0.835 & 0.999 & \textbf{0.859} & 0.024 & $10^{-15}$ \\
DecisionTree & 14 & 0.950 & 0.884 & 0.955 & \textbf{0.895} & 0.011 & $10^{-05}$ \\
ExtraTree & 11 & 0.915 & 0.844 & 0.924 & \textbf{0.861} & 0.017 & $10^{-04}$ \\
ExtraTrees & 26 & 0.985 & 0.828 & 0.991 & \textbf{0.853} & 0.025 & $10^{-16}$ \\
GradientBoosting & 58 & 0.930 & 0.822 & 0.926 & \textbf{0.840} & 0.018 & $10^{-17}$ \\
HistGradientBoosting & 52 & 0.961 & 0.820 & 0.963 & \textbf{0.839} & 0.019 & $10^{-19}$ \\
RandomForest & 18 & 0.992 & 0.855 & 0.997 & \textbf{0.881} & 0.026 & $10^{-11}$ \\
Total & 199 & 0.958 & 0.832 & 0.960 & \textbf{0.852} & 0.020 & $10^{-74}$ \\
\hline 
\end{tabular}
\end{center}
\end{table} %

\section{Why Does PDL Yield Improved Performance?}\label{sec:discussion}

The empirical results reveal that the PDL classifier significantly improves over the baseline methods. In this section, we elaborate on possible reasons for this improvement. 

\subsection{Combining Instance-based and Model-based Learning}

A distinguishing feature of PDL is a unique combination of (local) \emph{instance-based} learning and (global) \emph{model-based} learning. Like the well-known nearest-neighbor principle, a prediction for a new query is produced by other instances from the training set, namely the anchor points; yet, as opposed to NN, these instances are not restricted to nearby cases but can be located anywhere in the instance space. This becomes possible through the model-based component of PDL, namely the classifier $\gamma$, which is a global model that generalizes over the entire instance space. Broadly speaking, by constructing $\gamma$, the classifier learns how to transfer class information from one data point to another. 

Of course, there are other learning methods with similar characteristics. For example, instead of using a predefined distance function, the nearest neighbor method can be instantiated with a distance function $\delta$ that is learned on the training data. Metric learning typically proceeds from sets of similar instances (belonging to the same class) and dissimilar instances (belonging to different classes), and seeks to learn a function $\delta$ that keeps the distance low for the former while making it high for the latter \cite{globerson2005metric,bian_la13}. In a sense, this is indeed quite comparable to PDL, especially because both $\delta$ and $\gamma$ are two-place functions taking pairs of instances as input. Moreover, $\gamma$ could indeed also be seen as a kind of distance measure, if ``distance'' is defined in terms of ``probability of belonging to the same class''. Yet, PDL is arguably more flexible, because $\gamma$ is not required to satisfy properties of a distance or metric.   

\subsection{Simplification through Binary Reduction}

Another advantage of PDL is \emph{simplicity}: The original classification task is effectively reduced to a \emph{binary} problem, namely, to decide whether or not two instances share the same class label. This is comparable to binary decomposition techniques such as one-vs-rest and all-pairs \cite[p.202]{bishop2006pattern}, which reduce a single multinomial classification problem to several binary problems. Instead, PDL constructs a \emph{single} binary problem, although the total number of training examples produced essentially coincides for all methods (it is roughly quadratic in the size of the original data). In any case, binary problems are normally easier to solve, which explains the improved classification accuracy commonly reported for reduction techniques. In this regard, a decomposition can even be useful for methods that are able to handle multinomial problems right away (such as decision trees). 

\subsection{Error Reduction through Averaging}

Last but not least, by instantiating the global model for every anchor and collecting predictions from all of them, PDL benefits from a kind of ensemble effect and reduces error through \emph{averaging}. In particular, since prediction errors of individual anchors can be compensated by other anchors, PDL is able to reduce the variance of the prediction error. Again, this is somewhat comparable to the nearest-neighbor method. Given the model $\gamma$, the anchor predictions can even be considered as independent\footnote{Of course, this independence is lost if the anchor points are also part of the data used to train $\gamma$.}, which, under the simplified assumption of homoscedasticity, means that the prediction error is reduced by a factor of $1/\sqrt{A}$, with $A$ the number of anchors \cite[p.4]{wetzel2022btwin}. 

\begin{figure*}[!tb]
\centerline{\includegraphics[width=\columnwidth, trim=5 5 5 1, clip]{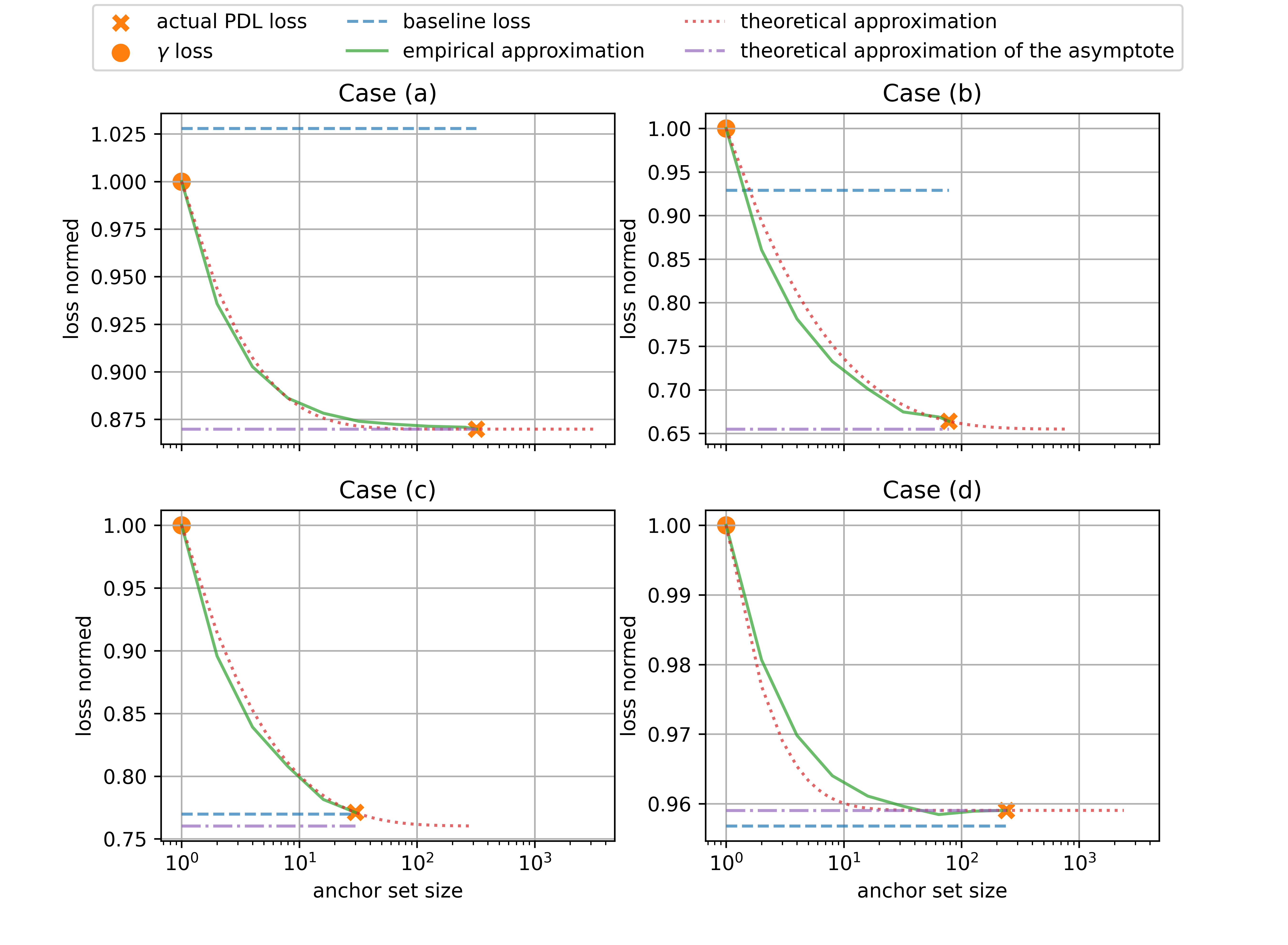}}
\caption{Effect of the anchor set size on PDC's loss relative to the baseline.}\label{fig:anchors_effect}
\end{figure*}

Even if these assumptions may not be completely satisfied, an expected improvement through averaging can clearly be observed in empirical studies. \cref{fig:anchors_effect} represents four cases encountered with four different datasets and DecisionTree as a baseline. 
We compare the loss of the baseline (baseline loss) with the actual PDL loss, i.e., the loss given all available anchors. 
The empirical approximation curve is meant to show how the loss depends on the number of anchor points. Its value at $A$ is produced by averaging the performance over randomly selected anchor subsets of size $A$. The curve goes from the average loss when only one anchor is used ($\gamma$ loss) until reaching the actual PDL loss.
The theoretical approximation curve is an optimal fit of a theoretical model to the empirical approximation, namely, the decrease of the error under the ideal assumption of independent prediction errors distributed normally with mean $\mu$ and standard deviation $\sigma$. As can be seen, even if this assumption may not fully hold, the two curves deviate but slightly. 

In case (a), the loss of the PDC's $\gamma$ estimator is better than the loss of the baseline model. As expected in this case, PDC is better than the baseline with any number of anchors.
In case (b), the baseline loss is between $\gamma$ loss and PDC's loss. With the theoretical approximation, we estimate how many anchors are enough to outperform the baseline.
In case (c), the baseline model is better than PDL. Nevertheless, the theoretical approximation allows us to estimate the additional anchors needed to outperform the baseline and the best reachable loss. It becomes less and less efficient to improve the score by adding more anchors. It might become more interesting, starting from a certain size, to work more on the base learner or the data quality. 
In case (d), the baseline model is even better than the approximated asymptote because learning the dual problem is more difficult. Adding more anchors is less likely to help.

\section{Conclusion}
Building on the concept of pairwise difference learning (PDL), we proposed the PDL classifier (PDC), a meta-learner able to reduce a multiclass classification problem into a binary problem. 
Our extensive empirical evaluation across 99 diverse datasets demonstrates that PDL consistently outperforms state-of-the-art machine learning models, resulting in improved F1 scores in a majority of cases. This highlights PDL's effectiveness in enhancing performance over baseline methods, facilitated through its straightforward integration via our Python package. To explain its strong performance, we also elaborated on several properties and features of PDC.

Future research directions include the exploration of instance (anchor) weighting through regularization or Shapley data importance \cite{ghorbani2019data} and interaction \cite{belaid2023optimizing}. 
Moreover, we plan to elaborate more closely on PDC's potential to quantify predictive uncertainty (cf.\ Section \ref{sec:uq})

In conclusion, PDL emerges as a practical solution for improving ML models, offering versatility and performance improvements across diverse applications. Its adaptability and robust performance make it a valuable addition to the ML toolkit, promising more accurate and reliable predictions in various domains.

\bibliographystyle{misc/splncs04} %
\bibliography{bib}

\end{document}